\documentclass[conference]{IEEEtran}
\IEEEoverridecommandlockouts
\usepackage{amsmath,amsfonts}
\usepackage{array}
\usepackage{textcomp}
\usepackage{stfloats}
\usepackage{url}
\usepackage{verbatim}
\usepackage{graphicx}
\usepackage{amssymb}
\usepackage{hyperref}
\usepackage{graphicx}
\usepackage{amsmath}
\usepackage{longtable}
\usepackage{algorithm} 
\usepackage{algpseudocode} 
\usepackage{mathrsfs}
\usepackage{subcaption}
\usepackage{mathtools}
\usepackage{pifont}
\usepackage{color}
\usepackage{lineno}
\usepackage{makecell} 
\usepackage{pdflscape}
\usepackage{adjustbox}
\usepackage[utf8]{inputenc}
\usepackage{tabularx}
\usepackage{blindtext}
\usepackage{longtable}
\usepackage{lscape}
\usepackage{setspace}
\usepackage{notoccite} 
\usepackage{lscape} 
\usepackage{mwe}
\usepackage{booktabs}
\usepackage{amsthm}

\theoremstyle{definition}

\theoremstyle{definition}

\newcommand{\RNum}[1]{\lowercase\expandafter{\romannumeral #1\relax}}
\newcommand{\RNumU}[1]{\uppercase\expandafter{\romannumeral #1\relax}}
\usepackage[numbers]{natbib}

\def\BibTeX{{\rm B\kern-.05em{\sc i\kern-.025em b}\kern-.08em
    T\kern-.1667em\lower.7ex\hbox{E}\kern-.125emX}}
\begin{document}

\title{Twin Restricted Kernel Machines for Multiview Classification\\
}

\author{
\IEEEauthorblockN{A. Quadir}
\IEEEauthorblockA{
\textit{Department of Mathematics} \\
\textit{Indian Institute of Technology Indore}\\
\textit{Indore, India}\\
mscphd2207141002@iiti.ac.in}
\and
\IEEEauthorblockN{M. Sajid}
\IEEEauthorblockA{
\textit{Department of Mathematics} \\
\textit{Indian Institute of Technology Indore}\\
\textit{Indore, India}\\
phd2101241003@iiti.ac.in }
\and
\IEEEauthorblockN{Mushir Akhtar}
\IEEEauthorblockA{
\textit{Department of Mathematics} \\
\textit{Indian Institute of Technology Indore}\\
\textit{Indore, India}\\
phd2101241004@iiti.ac.in }
\and
\IEEEauthorblockN{M. Tanveer}
\IEEEauthorblockA{
\textit{Department of Mathematics} \\
\textit{Indian Institute of Technology Indore}\\
\textit{Indore, India}\\
mtanveer@iiti.ac.in}
}



\maketitle

\begin{abstract}
Multi-view learning (MVL) is an emerging field in machine learning that focuses on improving generalization performance by leveraging complementary information from multiple perspectives or views. Various multi-view support vector machine (MvSVM) approaches have been developed, demonstrating significant success. Moreover, these models face challenges in effectively capturing decision boundaries in high-dimensional spaces using the kernel trick. They are also prone to errors and struggle with view inconsistencies, which are common in multi-view datasets. In this work, we introduce the multiview twin restricted kernel machine (TMvRKM), a novel model that integrates the strengths of kernel machines with the multiview framework, addressing key computational and generalization challenges associated with traditional kernel-based approaches. Unlike traditional methods that rely on solving large quadratic programming problems (QPPs), the proposed TMvRKM efficiently determines an optimal separating hyperplane through a regularized least squares approach, enhancing both computational efficiency and classification performance. The primal objective of TMvRKM includes a coupling term designed to balance errors across multiple views effectively. By integrating early and late fusion strategies, TMvRKM leverages the collective information from all views during training while remaining flexible to variations specific to individual views. The proposed TMvRKM model is rigorously tested on UCI, KEEL, and AwA benchmark datasets. Both experimental results and statistical analyses consistently highlight its exceptional generalization performance, outperforming baseline models in every scenario.
\end{abstract}

\begin{IEEEkeywords}
Restricted Boltzmann machines, Kernel methods, Restricted kernel machines, Multiview support vector machine.
\end{IEEEkeywords}

\section{Introduction}
\IEEEPARstart{M}{ultiview} learning (MVL) focuses on datasets that can be described using multiple distinct sets of features \cite{xu2016discriminatively}. Different views often offer complementary information to one another. Unlike conventional single-view learning, which relies on a single representation, MVL develops distinct learning models for different views and optimizes them jointly by utilizing the shared information among different perspectives of the same data. Traditional models often merge multiple views into a unified representation for a comprehensive description. However, this straightforward strategy of concatenating views not only results in the curse of dimensionality but also overlooks the unique statistical characteristics of each view. To date, a wide range of MVL models have been developed, which can be broadly divided into two categories: co-regularization-based models \cite{mao2014generalized, guan2015multi} and co-training-based models \cite{wang2010new, kumar2011co, wang2015unsupervised}.

Many multi-view classification approaches adopt a late fusion strategy, where information from different views is combined toward the end of the training process. A common example includes committee-based methods \cite{perrone1995networks}, where individual models are trained separately on each view, and their predictions are later aggregated, often using weighted combinations. The benefit of late fusion techniques lies in the flexibility of the individual submodels to independently adapt to the unique characteristics of each view, making them particularly effective when the data varies significantly across views. However, a key drawback is that since information is merged only at the final stage, each submodel has limited access to insights from other views during training. Early fusion methods integrate information from multiple views right from the beginning of the training process. A common example is feature concatenation, where data from different views are merged into a unified representation, as demonstrated by \citet{karevan2015black} work on temperature prediction using measurements from various cities. To capitalize on the strengths of both fusion techniques, some multi-view classification models strive to incorporate multi-view information early while preserving the flexibility to process each view separately.

Despite the noticeable differences among existing MVL models, they primarily reflect either the complementarity principle or the consensus principle in their learning approach \cite{xu2014large}. The consensus principle focuses on ensuring consistency across different views, indicating that each view carries shared information, which collectively contributes to a unified representation. Complementarity, in contrast, emphasizes that each view holds distinct information, meaning that each view provides specific, unique details. Effectively utilizing the relationships between different views can significantly enhance generalization performance \cite{tang2018learning}. There are two main approaches for handling multi-view data. The connection-based strategy is the first approach, which combines all the different views into one unified and enhanced representation. The second is the separation-based approach, where individual learning functions are assigned to each view, capitalizing on shared alignment features within each perspective. Notable algorithms in this category include SVM-2K \cite{farquhar2005two} and canonical correlation analysis (CCA) \cite{hardoon2004canonical}.

SVM-2K enhances the SVM framework by incorporating a co-regularization term into the margin maximization problem, allowing it to effectively leverage information from multi-view data for improved classification performance. Over the years, numerous multi-view classification models based on SVM have emerged, including multi-view twin SVM (MvTSVM) \cite{xie2015multi}, multi-view Laplacian SVM (MvLapSVM) \cite{sun2011multi}, and multi-view Laplacian TSVM (MvLapTSVM) \cite{xie2014multi}. Expanding on this concept, \citet{xu2022multi} introduced the multi-view learning framework with the privileged weighted TSVM (MPWTSVM). This approach builds a multi-view model that leverages the consensus principle and the K-Nearest Neighbor algorithm to extract intra-class information effectively. \citet{xie2020multi} introduced a manifold-preserving graph reduction (MPGR) technique to identify two sparse subsets from separate views and merge them through intersections, leading to the development of the MvLapLSTSVM model. \citet{ye2021multiview} proposed a robust double-sided twin SVM (MvRDTSVM) that improves classification resilience by incorporating \(L_1\)-norm distance metrics and bidirectional constraints. Additionally, they introduced a faster variant, MvFRDTSVM, to improve computational efficiency. Since then, a range of multi-view classification methods based on SVM have been consistently introduced, including multiview privileged SVM (PSVM-2V) \cite{tang2017multiview}, multi-view hypergraph regularized Lp norm least squares TSVM (MvHGLpLSTSVM) \cite{lu2024multi}, intuitionistic fuzzy multi-view SVM with Universum data (U-IFMvSVM) \cite{lou2024intuitionistic}, multiview learning with twin parametric margin SVM (MvTPMSVM) \cite{quadir2024multiview}, and multiview SVM with wave loss (Wave-MvSVM) \cite{quadir2024enhancing}. 

MvSVM and its variants demonstrate competitive accuracy and lower computational cost than numerous advanced methods. The restricted kernel machine (RKM) \cite{suykens2017deep, QUADIR2025108449, quadir2024one}, and MvRKM  \cite{houthuys2021tensor} are an innovative approach designed to integrate the strengths of kernel methods with neural networks. Building upon the foundation of LSSVM, it provides a structure that incorporates both visible and hidden units, resembling the architecture of the restricted Boltzmann machine (RBM) \cite{hinton2006fast}. Inspired by the strengths of RKM, we introduce the multiview twin restricted kernel machine (TMvRKM). This model combines the concepts of RKM with the advantages of twin methods, providing a more effective solution for classification tasks. TMvRKM utilizes the kernel function to transform the data into higher-dimensional spaces effectively, enabling better separation of complex patterns. However, the aforementioned SVM-based MVL methods involve solving large quadratic programming problems (QPPs), which can be computationally expensive, particularly for large-scale datasets. In contrast, the proposed TMvRKM model is designed to enhance efficiency, effectively mitigating the computational challenges associated with complex datasets. The primary objective of TMvRKM includes a coupling term aimed at reducing the combined error across all views. This approach effectively balances the benefits of both early and late fusion by integrating multi-view information during training while maintaining the adaptability of individual views. As a result, TMvRKM leverages shared insights from all perspectives while preserving the unique characteristics of each view, enhancing overall model performance. The following are the main highlights of this paper:
\begin{enumerate}
    \item In this paper, we propose a novel multiview twin restricted kernel machine (TMvRKM) model. TMvRKM model efficiently captures complementary information from different feature spaces, enhancing its classification accuracy.
    \item In the TMvRKM model, the kernel trick is used to transform the data into a higher-dimensional feature space, enabling more complex relationships to be captured during learning. Within this space, the algorithm identifies a hyperplane that optimally separates the training instances using a regularized least squares approach.
    \item Unlike traditional SVM-based MVL methods that rely on solving large quadratic programming problems (QPPs), the TMvRKM model is designed to optimize computational efficiency, making it more scalable for large-scale datasets.
    \item The TMvRKM model integrates the strengths of both early and late fusion by allowing each view to have its own regularization parameters. At the same time, the coupling term plays a crucial role in minimizing the overall error by ensuring that the combined effect of all error variables remains as small as possible. This design enables the model to effectively balance flexibility across views while maintaining consistency in error reduction.  
    \item To evaluate the generalization capability of the proposed TMvRKM model, we conduct experiments using datasets from UCI, KEEL, and AwA. The experimental findings demonstrate that TMvRKM consistently outperforms the baseline models, highlighting its effectiveness in multi-view learning tasks.
\end{enumerate}
 The general outline of this paper: Section \ref{Related-Work} gives a brief review of the baseline MvTSVM and MvRKM models. A comprehensive mathematical formulation of the TMvRKM model is provided in Section \ref{TMvRKM}. In Section \ref{Experiments and Results}, we discuss the experimental setup, results, and comparative analysis between the proposed and existing models.  In the end, we make conclusions in Section \ref{Conclusions-section}.

\section{Related Work}\label{Related-Work}
In this section, we briefly outline the mathematical formulations of MvTSVM and MvRKM. Consider \( V \) as the number of distinct views and the training set consists of \( K \) data points, denoted as \( \{ y^{(l)}_k, x^{[v]}_k \} \), where \( k = 1, \dots, K \) and \( l = 1, \dots, m \), for each view \( v = 1, \dots, V \). Here, \( x^{[v]}_k \in \mathbb{R}^{d[v]} \) represents the \( k \)-th input pattern in the \( v \)-th view, while \( y^{(l)}_k \in \{-1, +1\} \) represents the \( l \)-th output unit associated with the \( k \)-th label. Let \( \mathbf{A}^{[v]} \) and \( \mathbf{B}^{[v]} \) are the input matrices associated with the \( v \)-th view. These matrices capture the feature representations of the data for the respective view.

\subsection{Multiview twin support vector machine (MvTSVM)}
In the MvTSVM model \cite{xie2015multi}, four non-parallel hyperplanes need to be defined. Let \( \textbf{A}^A \) and \( \textbf{B}^A \) denote the samples corresponding to the \( +1 \) and \( -1 \) classes for the \( 1^{st} \) view, respectively, while \( \textbf{A}^B \) and \( \textbf{B}^B \) correspond to the \( +1 \) and \( -1 \) class samples for \( 2^{nd} \) view. To define the data matrices, let \( A_1 = [\textbf{A}^A, e_1] \), \( A_2 = [\textbf{B}^A, e_2] \), \( B_1 = [\textbf{A}^B, e_1] \), and \( B_2 = [\textbf{B}^B, e_2] \), where \( e_1 \in \mathbb{R}^{m_1} \) and \( e_2 \in \mathbb{R}^{m_2} \) are vectors of ones. Additionally, let \( v_1 = [w_1^A; b_1^A] \), \( v_2 = [w_1^B; b_1^B] \), \( u_1 = [w_2^A; b_2^A] \), and \( u_2 = [w_2^B; b_2^B] \). The hyperplanes of MvTSVM model can be found by solving the following QPP:
\begin{align}
\label{eq:4}
\underset{  \eta_1~ \xi_1,~ \xi_2,~v_{1},~ v_{2}}{\min}  \hspace{0.1cm}~&\frac{1}{2}\|A_2v_2\|^2 + \frac{1}{2}\|A_1v_1\|^2 +C_{1}e_2^T\xi_{1} + C_{2}e_2^T\xi_{2}   \nonumber \\
& + D_1e_1^T\eta_{1} \nonumber \\
 \text { s.t. }\hspace{0.1cm}  & \lvert A_1v_1 - A_2v_2 \rvert \leq \epsilon + \eta_1, \nonumber \\
 & -B_1v_1 \geq e_2 - \xi_1, \nonumber \\
 & -B_2v_2 \geq e_2 - \xi_2, \nonumber \\
 & \xi_{1} \geq 0, \xi_{2} \geq 0, \eta_{1}\geq 0, 
\end{align}
and
\begin{align}
\label{eq:5}
\underset{ \eta_2,~ \xi_3,~ \xi_4,~ u_{1},~ u_{2}}{\min}  \hspace{0.1cm}~&\frac{1}{2}\|B_2u_2\|^2+\frac{1}{2}\|B_1u_1\|^2+C_{3}e_1^T\xi_{3} + C_{4}e_1^T\xi_{4}  \nonumber \\
& + D_2e_2^T\eta_{2} \nonumber \\
 \text { s.t. }\hspace{0.1cm}  & \lvert B_1u_1 - B_2u_2 \rvert \leq \epsilon + \eta_2, \nonumber \\
 & A_1u_1 \geq e_1 - \xi_3, \nonumber \\
 & A_2u_2 \geq e_1 - \xi_4, \nonumber \\
 & \xi_{3} \geq 0, \xi_{4} \geq 0, \eta_{2}\geq 0, 
\end{align}
where $C_1$, $C_2$, $C_3$, $C_4$, $D_1$, $D_2$ are tunable parameters, and $\epsilon > 0$ is an insensitive parameter; $\xi_1$, $\xi_2$, $\xi_3$, $\xi_4$, $\eta_1$, and $\eta_2$ are slack variables, respectively.

\subsection{Multi-view RKM classification}
A training set of \(K\) data points \(\{(\mathbf{x}_k^{[v]}, y_k)\}_{k=1}^{N}\) for each view \(v = 1, \dots, V\) and a set of \(V\) views are given. The objective function of MvRKM  \cite{houthuys2021tensor} model is defined as follows:
\begin{align}
\label{MvRKM1}
    \mathcal{J} = &\sum_{v=1}^{V} \sum_{k=1}^{N} \left( 1 - \left( \varphi^{[v]}(\mathbf{x}_k^{[v]})^T \mathbf{w}^{[v]} + b \right) y_k \right) h_k \nonumber \\ 
    & + \eta \sum_{v=1}^{V} \mathbf{w}^{[v]T} \mathbf{w}^{[v]} - \frac{\lambda}{2} \sum_{k=1}^{N} h_k^2,
\end{align}
where \(\varphi^{[v]}: \mathbb{R}^{d[v]} \rightarrow \mathbb{R}^{d[v]}\) represents the view-specific feature maps, \(\lambda\) and \(\eta~\) are positive terms, and \(h\) denotes the high-dimensional feature space. Instead of directly working with the feature maps, we employ a kernel function \(K^{[v]}: \mathbb{R}^{d[v]} \times \mathbb{R}^{d[v]} \rightarrow \mathbb{R}\).

The solution to the optimization problem \eqref{MvRKM1} is given by:
\begin{align}
    \begin{bmatrix}
\mathbf{V}_N & \frac{1}{\eta} \sum_{v=1}^{V} \Phi^{[v]} + \lambda \mathbf{I}_N \\
0  & \mathbf{V}_N^T
\end{bmatrix}
\begin{bmatrix}
b \\
\mathbf{y} \odot \mathbf{h}
\end{bmatrix}
= 
\begin{bmatrix}
\mathbf{V} \mathbf{y} \\
0
\end{bmatrix},
\end{align}
where \( \Phi^{[v]} \) denotes the kernel matrix corresponding to view \( v \) and \( \mathbf{V}_N = V \times e \in \mathbb{R}^{N \times 1} \).

\section{Twin Multiview Restricted Kernel Machine (TMvRKM)}
\label{TMvRKM}
This section delves into the proposed twin multiview restricted kernel machine (TMvRKM). To start, we introduce the mathematical foundation that defines TMvRKM. Unlike traditional kernel methods, TMvRKM adopts a novel strategy by incorporating both visible and hidden variables in its formulation. This dual-variable integration parallels the energy function employed in restricted Boltzmann machines (RBM) \cite{hinton2006fast}, creating a conceptual link between kernel-based methodologies and RBM. This connection not only highlights the versatility of TMvRKM but also demonstrates its ability to leverage the strengths of both approaches for robust classification. The function \( \phi: \mathbf{x}_i \to \phi(\mathbf{x}_i) \) transforms the training samples from the input space into a high-dimensional feature space, facilitating enhanced representation during both the training and prediction stages. The formulation of TMvRKM extends the RKM classification framework to accommodate data from multiple views. These views are interconnected through shared hidden features, enabling the model to capture and leverage underlying dependencies across the views for improved classification performance.

The formulation of TMvRKM for the first hyperplane is given as follows:
\begin{align}
\label{TMvRKM:1}
    & J_1 =  \underset{w^{[v]}}{\min}  \sum_{v=1}^V e_2^T(B^{[v]}w^{[v]} + e_1b_1^{[v]}) + \frac{\eta_1}{2} \sum_{v=1}^V w^{[v]^T}w^{[v]} \nonumber  \\
    & ~~~~ +  \frac{1}{2\lambda_1} \sum_{v=1}^V \xi_1^{[v]^T}\xi_1^{[v]} \nonumber \\
    & ~~\text{s.t.} ~~  A^{[v]}w^{[v]} + e_1b_1^{[v]} = e_1 - \xi_1^{[v]},
\end{align}
where $\xi_1$ is a slack variable, $e_1$ and $e_2$ are the vector of ones of suitable dimensions, and $w$ is the interconnection matrix. The TMvRKM establishes an upper bound for $J_1$ by incorporating the hidden layer representations $h_1$ as follows:
\begin{align}
\label{TMvRKM:2}
    \frac{1}{2\eta_1}\xi_1\xi_1^T  \geq \xi_1^Th_1 - \frac{\eta_1}{2} h_1^Th_1.
\end{align}
From \eqref{TMvRKM:1} and \eqref{TMvRKM:2}, we obtain:
\begin{align}
    L_1 & = \frac{\eta_1}{2} \sum_{v=1}^V w^{[v]^T}w^{[v]} + \sum_{v=1}^V e_2^T (B^{[v]}w^{[v]} + eb_1^{[v]}) \nonumber \\
    & + \sum_{v=1}^V(e-A^{[v]}w^{[v]} - eb_1^{[v]})^Th_1 - \frac{\lambda_1}{2}h_1^T h_1.
\end{align}

The points where the objective function reaches its stationary values can be determined as follows:
\begin{align}
    & \frac{\partial L_1}{\partial w^{[v]}} = \eta_1w^{[v]} + B^{[v]^T}e_2 - A^{[v]^T}h_1 = 0, \label{P:4} \\
   & \frac{\partial L_1}{\partial h_1} = \sum_{v=1}^V (e-A^{[v]}w^{[v]} -eb_1^{[v]}) -\lambda_1 h_1 =0, \label{P:5} \\
   & \frac{\partial L_1}{\partial b} = e_2^T e_2 - e^Th_1 = 0.
\end{align}
From \eqref{P:4}, we obtain:
\begin{align}
\label{P:7}
    w^{[v]} = \frac{1}{\eta_1} \left [A^{[v]^T}h_1 - B^{[v]^T}e_2 \right ].
\end{align}
Substituting $w^{[v]}$ from \eqref{P:7} in \eqref{P:5}, we get:
\begin{align}
\label{P:8}
    V_e -V_N b_1 -\frac{1}{\eta_1} \sum_{v=1}^V A^{[v]} \left[ A^{[v]^T} h_1 - B^{[v]^T}e_2 \right] - \lambda_1 h_1 = 0.
\end{align}
The kernel matrix \(\mathcal{K}\) is constructed using a kernel function that captures the inner product in the high-dimensional feature space. It is defined as \(\mathcal{K}(A, C) = \phi(A)\phi(C)^T\), where the mapping \(\phi\) transforms the data into the feature space. As a result, \eqref{P:8} can be expressed as:
\begin{align}
 V_e - V_N b_1 -\frac{1}{\eta_1} \sum_{v=1}^V \mathcal{K}(A^{[v]}, A^{[v]})h_1 - & \frac{1}{\eta_1}\sum_{v=1}^V \mathcal{K}(A^{[v]}, B^{[v]})e_2 \nonumber \\
 & - \lambda_1h_1 = 0.
\end{align}
By eliminating the weight variables, the problem simplifies to a linear system of equations, which can be obtained as follows:
\begin{align}
    \left[\begin{array}{c|c } 
	\frac{1}{\eta_1}\sum_{v=1}^V \mathcal{K}(A^{[v]}, A^{[v]}) + \lambda_1 I & V  \\  
	\hline 
	 e_2^T    & 0  
\end{array}\right]
    \begin{bmatrix}
    h_1 \\ b_1
    \end{bmatrix} \nonumber \\ =  \begin{bmatrix} V_e - \frac{1}{\eta_1} \sum_{v=1}^V \mathcal{K}(A^{[v]}, B^{[v]^T})e_2 \\   e_2^Te_2   \end{bmatrix}.
\end{align}

Similarly, the formulation of TMvRKM for the second hyperplane is defined as follows:
\begin{align}
\label{TMvRKM:02}
     J_2 = & \underset{w}{\min} \sum_{v=1}^V e_1^T(A^{[v]}u^{[v]} + e_2b_2^{[v]}) + \frac{\eta_2}{2} \sum_{v=1}^V u^{[v]^T}u^{[v]} \nonumber \\
    & +  \frac{1}{2\lambda_2} \sum_{v=1}^V \xi_2^{[v]^T}\xi_2^{[v]} \nonumber \\
    &\hspace{-0.2cm} \text{s.t.} ~~ B^{[v]}u^{[v]} + e_2b_2^{[v]} = \xi_2^{[v]} - e_2,
\end{align}
where \(\xi_2\) represents the slack variables and \(u\) denotes the interconnection matrix. The Lagrangian function corresponding to the problem \eqref{TMvRKM:02} can be formulated as:
\begin{align}
    L_2 = & \frac{\eta_2}{2} \sum_{v =1}^V u^{[v]^T}u^{[v]} + \sum_{v=1}^V e_2^T(A^{[v]} u^{[v]} + eb_2^{[v]}) \nonumber \\
    & + \sum_{v=1}^V(e_2 + B^{[v]} u^{[v]} + eb_2^{[v]})^Th_2 - \frac{\lambda_2}{2} h_2^Th_2.
\end{align}
By computing the partial derivatives of the objective function, we can determine the stationary points as follows:
\begin{align}
    & \frac{\partial L_2}{\partial u^{[v]}} = \eta_2u^{[v]} + A^{[v]^T}e_2 + B^{[v]^T}h_1 = 0, \\
   & \frac{\partial L_2}{\partial h_2} = \sum_{v=1}^V (e_2 + B^{[v]}u^{[v]} + eb_2^{[v]}) -\lambda_2 h_2 =0, \\
   & \frac{\partial L_2}{\partial b} = e_1^T e_1 - e^Th_2 = 0.
\end{align}

By removing the weight variables, the problem simplifies into the following linear form:
\begin{align}
    \left[\begin{array}{c|c } 
	\frac{1}{\eta_2}\sum_{v=1}^V \mathcal{K}(B^{[v]}, B^{[v]}) + \lambda_2 I & V  \\  
	\hline 
	 e_1^T    & 0  
\end{array}\right]
    \begin{bmatrix}
    h_2 \\ b_2
    \end{bmatrix}  \nonumber \\=  \begin{bmatrix} V_e - \frac{1}{\eta_2} \sum_{v=1}^V \mathcal{K}(B^{[v]}, A^{[v]^T})e_1 \\   e_1^Te_1   \end{bmatrix}.
\end{align}
Once the optimal values of $h_1$ ($b_1$) and $h_2$ ($b_2$) are calculated for the $+1$ and $-1$ class, respectively. To predict the label of a new sample \( x \), the following decision function can be used as follows:
\begin{align}
\label{eq:22}
    \text{class}(x) =  sign\left(f_1(x) + f_2(x)\right),
\end{align}
where
\begin{align}
    f_1(x) = \frac{1}{\eta_1} \left [\sum_{v=1}^V\mathcal{K}(x^{[v]},A^{[v]})h_1 - \sum_{v=1}^V\mathcal{K}(x^{[v]},B^{[v]})e_2 \right ]
\end{align}
and
\begin{align}
    f_2(x) = \frac{1}{\eta_2} \left [\sum_{v=1}^V\mathcal{K}(x^{[v]},B^{[v]})h_2 + \sum_{v=1}^V\mathcal{K}(x^{[v]},A^{[v]})e_1 \right ].
\end{align}

\section{Experiments, Results and Discussion}\label{Experiments and Results}
To evaluate the performance of the proposed TMvRKM model, we conduct experiments using publicly available benchmark datasets. These include $27$ real-world datasets from UCI \cite{dua2017uci} and KEEL \cite{derrac2015keel}, as well as 27 Animal with Attributes (AwA) \cite{tang2017multiview} binary classification datasets.

\subsection{Experimental setup}
In this subsection, we perform experiments by implementing 5-fold cross-validation combined with a grid search method to fine-tune the model's hyperparameters. The dataset is divided randomly in a $70:30$ proportion, where $70\%$ is allocated for training the model, and the remaining $30\%$ is used for testing. In all experiments \( \mathcal{K}(x_i, x_j) = e^{-\frac{\|x_i - x_j\|^2}{2\sigma^2}} \) is employed as a kernel function with the kernel parameter \( \sigma \) chosen from the set \( \{2^{-5}, 2^{-4}, \dots, 2^{4}, 2^{5}\} \). To minimize computational complexity, we assign equal penalty parameters, namely \( \eta_1 = \eta_2 \) and \( \lambda_1 = \lambda_2 \), selecting them from the same range \( \{10^{-5}, 10^{-4}, \dots, 10^{4}, 10^{5}\} \). This strategy streamlines the model while maintaining effective performance optimization. The experiments are conducted in Python 3.11 on Windows 11 running on a computer with system configuration Intel® Xeon® Gold 6226R CPU and 128 GB of RAM.

\begin{table*}[ht!]
\centering
    \caption{Classification accuracy (Acc) of the proposed TMvRKM  with the existing models on real-world datasets, i.e., KEEL and UCI.}
    \label{Average ACC and average rank for UCI and KEEL datasets}
    \resizebox{0.8\linewidth}{!}{
\begin{tabular}{lcccccc}
\hline
Model $\rightarrow$ &  SVM2K \cite{farquhar2005two} & MvTSVM \cite{xie2015multi} & MVLDM \cite{hu2024multiview} & MvTPMSVM \cite{quadir2024multiview} & MvRKM \cite{houthuys2021tensor} & TMvRKM$^{\dagger}$ \\ \hline
Dataset $\downarrow$ &  Acc (\%) & Acc (\%)  &  Acc (\%)  &  Acc (\%)  &  Acc (\%)  &  Acc (\%) \\ \hline
aus & 87.02 & 71.15 & 71.98 & 83.57 & 84.13 & 87.98 \\
bank & 80.74 & 71.86 & 73.67 & 89.31 & 89.09 & 88.21 \\
breast\_cancer\_wisc\_diag & 95.49 & 88.6 & 93.15 & 96.47 & 98.25 & 97.62 \\
breast\_cancer\_wisc\_prog & 58.33 & 58.33 & 71.17 & 72.88 & 75 & 73.68 \\
breast\_cancer & 62.45 & 55.58 & 70 & 78.82 & 76.74 & 77.33 \\
breast\_cancer\_wisc & 90.04 & 81.43 & 75 & 95.69 & 96.67 & 96.74 \\
brwisconsin & 97.56 & 61.95 & 95.59 & 96.57 & 96.1 & 96.83 \\
bupa or liver-disorders & 54.8 & 42.31 & 55.34 & 72.34 & 66.35 & 96.1 \\
checkerboard\_Data & 87.02 & 43.75 & 84.06 & 86.47 & 87.98 & 88.46 \\
chess\_krvkp & 80.45 & 82.35 & 97.7 & 94.99 & 99.37 & 84.13 \\
cleve & 80 & 75.56 & 84.27 & 77.53 & 81.11 & 86.03 \\
cmc & 64.25 & 55.88 & 74.38 & 64.41 & 71.49 & 80 \\
cylinder\_bands & 68.18 & 60.39 & 71.9 & 75.16 & 77.92 & 75.32 \\
conn\_bench\_sonar\_mines\_rocks & 80.95 & 46.03 & 75.81 & 82.26 & 80.95 & 82.6 \\
fertility & 75.25 & 75 & 86.67 & 86.67 & 85.42 & 86.89 \\
hepatitis & 80.85 & 78.72 & 78.26 & 91.3 & 82.98 & 80.85 \\
hill\_valley & 60.98 & 53.3 & 56.2 & 67.84 & 68.96 & 69.27 \\
mammographic & 80.27 & 77.06 & 83.33 & 81.25 & 82.01 & 91.51 \\
monks\_3 & 80.24 & 76.11 & 96.39 & 93.98 & 95.21 & 96.62 \\
new-thyroid1 & 78.46 & 82.31 & 95.31 & 100 & 98.46 & 92.31 \\
oocytes\_trisopterus\_nucleus\_2f & 78.83 & 58.39 & 82.05 & 73.63 & 86.86 & 86.98 \\
oocytes\_merluccius\_nucleus\_4d & 74.27 & 64.82 & 75.16 & 76.89 & 76.06 & 77.43 \\
parkinsons & 71.19 & 71.19 & 93.1 & 90.78 & 91.53 & 91.53 \\
pima & 76.19 & 73.33 & 69.13 & 70 & 76.19 & 70.13 \\
pittsburg\_bridges\_T\_OR\_D & 70.85 & 65.85 & 90 & 86.67 & 89.42 & 91.23 \\
ripley & 89.07 & 80.67 & 89.07 & 87.2 & 87.93 & 89.47 \\
planning & 65.85 & 63.64 & 68.52 & 83.33 & 76.36 & 76.36 \\
 \hline
Average Acc & 76.65 & 67.24 & 79.9 & 83.56 & 84.39 & \textbf{85.62} \\ \hline 
Average Rank & 4.22 & 5.7 & 3.74 & 3.06 & 2.48 & \textbf{1.8} \\ \hline
\multicolumn{7}{l}{$^{\dagger}$ represents the proposed model.}
\end{tabular}}
\end{table*}

\subsection{Results and discussions on real-world datasets i.e. UCI and KEEL}
This section provides an in-depth comparison of the performance of the proposed TMvRKM model against well-known baseline models, including SVM2K \cite{farquhar2005two}, MvTSVM \cite{xie2015multi}, MVLDM \cite{hu2024multiview}, MvTPMSVM \cite{quadir2024multiview}, and MvRKM \cite{houthuys2021tensor}. The evaluation is conducted using $27$ benchmark datasets from UCI \cite{dua2017uci} and KEEL \cite{derrac2015keel}. Our study encompasses a range of scenarios and includes thorough statistical analyses. Since the UCI and KEEL datasets do not naturally possess multiview characteristics, we create them artificially by designating the original dataset as view \( A \). Additionally, we derive a second view, \( B \), by extracting $95\%$ of the principal components from the original data, thereby enabling a multiview framework for evaluation.

Table \ref{Average ACC and average rank for UCI and KEEL datasets} presents the experimental results for the proposed TMvRKM model in comparison with the baseline SVM-2K, MvTSVM, MVLDM, MvTPMSVM, and MvRKM models. TMvRKM model reaches an average Acc of $85.62\%$. In comparison, the average Acc values for the SVM-2K, MvTSVM, MVLDM, MvTPMSVM, and MvRKM models are $76.65\%$, $67.24\%$, $79.90\%$, $83.56\%$, and $84.39\%$, respectively. The top accuracy achieved by our proposed TMvRKM model demonstrates its strong generalization capabilities. Since the average ACC can be influenced by exceptional performance on a single dataset, potentially compensating for weaker results on others, it may not always offer an impartial evaluation. To address this, we adopt a ranking method to evaluate the models' effectiveness. In this approach, each classifier is assigned a rank based on its performance, with the best-performing model receiving the lowest rank and the weakest one the highest. For \( l \) models evaluated across \( K \) datasets, \( r_i^j \) denotes the rank of the \( j \)-th model on the \( i \)-th dataset. The average rank for the \( j \)-th model is calculated as \( R_j = \frac{1}{K} \sum_{i=1}^K r_i^j \). The average ranks achieved by SVM-2K, MvTSVM, MVLDM, MvTPMSVM, MvRKM, and the proposed TMvRKM models are $4.22$, $5.70$, $3.74$, $3.06$, $2.48$, and $1.80$, respectively. The results show that the proposed TMvRKM model surpasses the baseline models, attaining the highest average rank. Consequently, the generalization capability of TMvRKM exceeds that of its counterparts. To evaluate the significance of the results, we conduct statistical tests to determine if the differences in performance between the models are meaningful. Specifically, we apply the Friedman test \cite{demvsar2006statistical}, which compares the average ranks assigned to each model in order to detect any significant performance disparities. The null hypothesis posits that the average ranks of all models are identical, implying that there are no significant differences in their performance. With \( (l - 1) \) degrees of freedom (d.o.f.), the Friedman test statistic has a chi-squared distribution, represented as \( \chi^2_F \). The test statistic is computed as: \( \chi^2_F = \frac{12K}{l(l+1)} \left[\sum_j R_j^2 - \frac{l(l+1)^2}{4}\right] \), where \( K \) represents the number of datasets, \( l \) the number of models, and \( R_j \) the average rank of the \( j \)-th model. Additionally, the Friedman statistic \( F_F \) can be determined as: \( F_F = \frac{(K - 1)\chi^2_F}{K(l - 1) - \chi^2_F} \), with degrees of freedom for the \( F \)-distribution being \( ((k - 1), (K - 1)(l - 1)) \). At a $5\%$ significance level, the calculated values are as follows: \( \chi^2_F = 73.5943 \) and \( F_F = 31.1608 \) for \( l = 6 \) models and \( K = 27 \) datasets. According to the \( F \)-distribution table, the critical value is \( F_F(5, 130) = 2.2839 \). Since \( 31.1608 > 2.2839 \), we reject the null hypothesis, confirming the existence of significant differences among the models. We use the Nemenyi post hoc test to investigate pairwise performance differences further. This test calculates the critical difference ($C.D.$) to identify significant differences in model performance. $C.D.$ is calculated as follows: $C.D. = q_\alpha \times \sqrt{l(l+1)/6K}$, where \( q_\alpha \) is the critical value derived from the distribution table of the Nemenyi test. At a $5\%$ significance level, \( q_\alpha = 2.850 \), and the $C.D.$ is determined as \( 1.4788 \). The differences in average ranks between the TMvRKM model and the existing models are as follows: 2.42, 3.70, 1.94, 1.26, and 0.68 for SVM-2K, MvTSVM, MVLDM, MvTPMSVM, and MvRKM, respectively. The rank differences between TMvRKM and SVM-2K, MvTSVM, and MVLDM surpass the critical difference threshold, indicating significant distinctions between the TMvRKM model and these baselines. Although the differences with MvTPMSVM and MvRKM do not surpass the critical difference, the TMvRKM model still outperforms these models in terms of average rank. These results demonstrate that the TMvRKM model outperforms its counterparts.

\begin{table*}[ht!]
\centering
    \caption{Classification accuracy (Acc) of the proposed TMvRKM with existing models across AwA datasets.}
    \label{Average ACC and average rank for AwA datasets}
    \resizebox{0.8\linewidth}{!}{
\begin{tabular}{lcccccc}
\hline
Model $\rightarrow$ &  SVM2K \cite{farquhar2005two} & MvTSVM \cite{xie2015multi} & MVLDM \cite{hu2024multiview} & MvTPMSVM \cite{quadir2024multiview} & MvRKM \cite{houthuys2021tensor} & TMvRKM$^{\dagger}$ \\ \hline
Dataset $\downarrow$ &  Acc (\%) & Acc (\%)  &  Acc (\%)  &  Acc (\%)  &  Acc (\%)  &  Acc (\%) \\ \hline
Chimpanzee vs   Leopard & 80.11 & 46.53 & 68.75 & 74.17 & 67.36 & 70.83 \\
Chimpanzee vs Persian cat & 70.86 & 50 & 86.11 & 75 & 65.97 & 87.36 \\
Giant panda vs Humpback whale & 93.06 & 46.53 & 93.75 & 85.83 & 88.19 & 93.11 \\
Giant panda vs Persian cat & 81.81 & 52.08 & 66.67 & 80 & 72.22 & 78.47 \\
Giant panda vs Raccoon & 80.19 & 52.78 & 64.58 & 68.33 & 70.83 & 80.47 \\
Hippopotamus vs Raccoon & 78.47 & 75.14 & 75.69 & 71.67 & 73.61 & 78.33 \\
Hippopotamus vs Rat & 75.33 & 75.83 & 64.58 & 68.33 & 68.75 & 68.75 \\
Humpback whale vs Raccoon & 85.67 & 80.69 & 83.33 & 80 & 79.86 & 86.39 \\
Humpback whale vs Rat & 82.28 & 80.31 & 77.78 & 80 & 78.47 & 83.08 \\
Humpback whale vs Seal & 76.39 & 72.08 & 78.47 & 77.5 & 72.22 & 79.25 \\
Leopard vs Hippopotamus & 78.17 & 50.69 & 75 & 74.17 & 70.14 & 75 \\
Leopard vs Persian cat & 82.19 & 79.31 & 80.56 & 81.67 & 80.56 & 83.33 \\
Leopard vs Humpback whale & 90.75 & 79.31 & 89.58 & 87.5 & 81.94 & 87.5 \\
Leopard vs Pig & 75 & 61.39 & 68.75 & 70 & 69.44 & 75.42 \\
Leopard vs Rat & 76.42 & 68.61 & 65.28 & 76.67 & 75.69 & 79.86 \\
Leopard vs Seal & 80.42 & 63.47 & 81.25 & 71.67 & 76.39 & 81.31 \\
Persian cat vs Humpback whale & 71.67 & 71.39 & 85.42 & 75 & 73.61 & 75 \\
Persian cat vs Hippopotamus & 76.81 & 76.53 & 75.69 & 80.83 & 75 & 75.69 \\
Persian cat vs Pig & 70 & 69.31 & 69.44 & 77.5 & 65.28 & 76.67 \\
Persian cat vs Raccoon & 82.64 & 79.31 & 65.97 & 72.5 & 75 & 82.83 \\
Persian cat vs Rat & 60.44 & 64.17 & 56.94 & 48.33 & 59.72 & 60.56 \\
Pig vs Humpback whale & 80.19 & 77.92 & 88.89 & 88.33 & 79.86 & 81.25 \\
Pig vs Raccoon & 71.69 & 69.31 & 62.5 & 57.5 & 61.11 & 71.69 \\
Pig vs Rat & 71.53 & 68.61 & 64.58 & 59.33 & 65.97 & 64.58 \\
Raccoon vs Rat & 62.22 & 61.89 & 65.28 & 72.5 & 65.28 & 70.14 \\
Raccoon vs Seal & 90.28 & 75.39 & 75.69 & 74.17 & 79.86 & 90.25 \\
Rat vs Seal & 70.86 & 65.17 & 69.87 & 70 & 65.97 & 71.22 \\ \hline
Average Acc & 77.61 & 67.18 & 74.09 & 74.02 & 72.53 & \textbf{78.09} \\ \hline
Average Rank & 2.35 & 4.78 & 3.72 & 3.67 & 4.46 & \textbf{2.02} \\ \hline
\multicolumn{7}{l}{$^{\dagger}$ represents the proposed model.}
\end{tabular}}
\end{table*}

\subsection{Results and discussions on AwA datasets}
To evaluate the performance of the proposed TMvRKM model, we perform experiments using the AwA dataset \cite{tang2017multiview}, which contains a total of $30,475$ images representing $50$ distinct animal species. Each image is accompanied by six distinct feature representations that have been pre-extracted. For the experiments, we select 10 animal classes based on Tang's work \cite{tang2017multiview}, which include giant panda, chimpanzee, humpback whale, Persian cat, leopard, pig, hippopotamus, rat, raccoon, and seal. These selected classes are organized into $27$ binary classification tasks using a one-vs-one approach.

Table \ref{Average ACC and average rank for AwA datasets} presents the Acc of the proposed TMvRKM model alongside the existing models SVM-2K, MvTSVM, MVLDM, MvTPMSVM, and MvRKM. The TMvRKM model consistently outperforms the existing models in terms of generalization performance across most datasets. Specifically, the TMvRKM model achieves an average Acc of $78.09\%$, whereas the baseline models achieve average Acc values of $77.61\%$ for SVM-2K, $67.18\%$ for MvTSVM, $74.09\%$ for MVLDM, $74.02\%$ for MvTPMSVM, and $72.53\%$ for MvRKM. The table also shows the average ranks based on the Acc values, where the TMvRKM model secures the lowest rank. To verify the statistical significance of these results, we apply the Friedman test and then conduct a Nemenyi post hoc analysis for further comparison. At \( \alpha = 0.05 \), the test yields \( \chi^2_F = 47.444 \), \( F_F = 14.0874 \), and a critical value of \( F_F(5,130) = 2.2839 \). Since \( F_F(5,130) < F_F \), the null hypothesis is rejected, indicating significant performance differences. The Nemenyi post hoc test calculates the $C.D.$ as $1.4788$, and differences in average ranks between the TMvRKM model and the existing models MvTSVM, MVLDM, MvTPMSVM, and MvRKM, with rank differences of $2.76$, $1.70$, $1.67$, and $2.44$, respectively. However, the difference between TMvRKM and SVM-2K is $0.33$, which does not exceed the $C.D.$ Despite this, the TMvRKM model still achieves a superior average rank compared to SVM-2K. These results clearly demonstrate that the TMvRKM model outperforms all baseline models except for SVM-2K, with a higher average rank overall, confirming its superior performance.

\subsection{Sensitivity Analyses}

\begin{figure*}[ht!]
\begin{minipage}{.246\linewidth}
\centering
\subfloat[breast\_cancer\_wisc\_prog]{\label{1a}\includegraphics[scale=0.22]{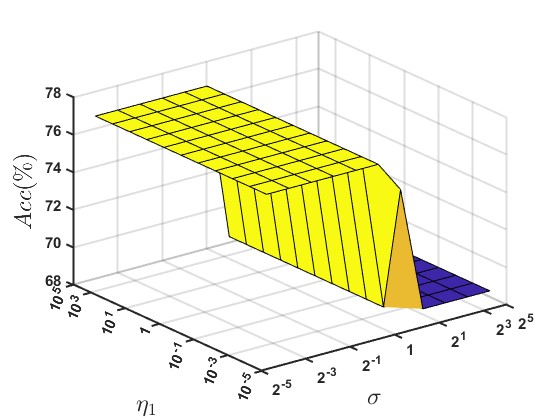}}
\end{minipage}
\begin{minipage}{.246\linewidth}
\centering
\subfloat[breast\_cancer\_wisc]{\label{1b}\includegraphics[scale=0.22]{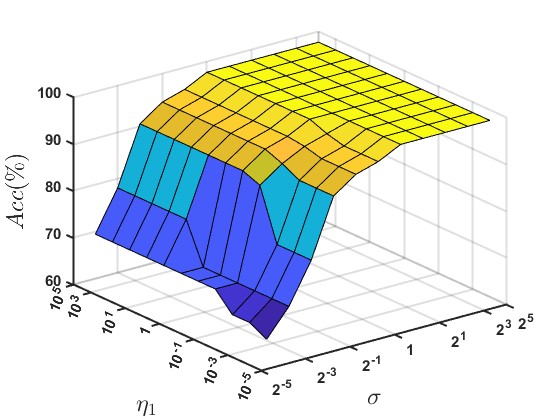}}
\end{minipage}
\begin{minipage}{.246\linewidth}
\centering
\subfloat[brwisconsin]{\label{1c}\includegraphics[scale=0.22]{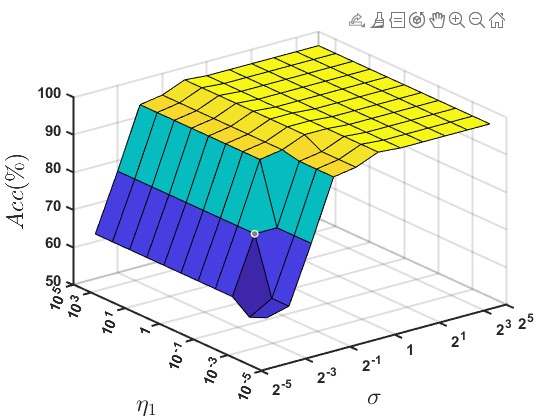}}
\end{minipage}
\begin{minipage}{.246\linewidth}
\centering
\subfloat[checkerboard\_Data]{\label{1d}\includegraphics[scale=0.22]{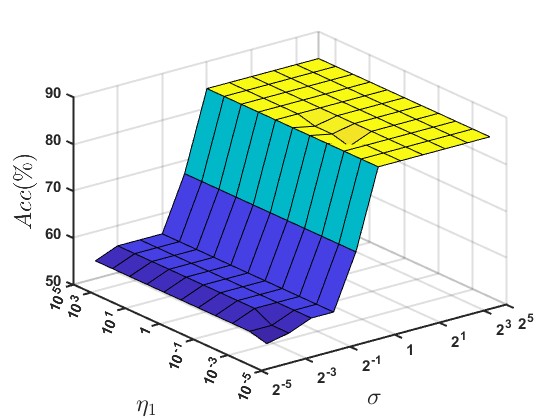}}
\end{minipage}
\caption{Effect of varying parameters $\eta_1$ and $\sigma$ on the Acc values of the proposed TMvRKM model.}
\label{Effect of parameters eta and gamma}
\end{figure*}

Assessing the robustness of the proposed TMvRKM model requires analyzing its sensitivity to the hyperparameters \(\eta_1\) and \(\sigma\). This detailed evaluation will help identify the best parameter settings that maximize predictive accuracy and strengthen the model's capacity to generalize effectively to new, unseen data. Fig \ref{Effect of parameters eta and gamma} highlights notable variations in Acc across different values of \(\eta_1\) and \(\sigma\), emphasizing the impact of these hyperparameters on model performance. As shown in Figs. \ref{1b} and \ref{1c}, the best performance is achieved when \(\eta_1\) and \(\sigma\) are within the ranges of \(10^{-1}\) to \(10^{5}\) and \(10^{0}\) to \(10^{5}\), respectively. Furthermore, Fig. \ref{1a} shows that the model achieves peak accuracy when \(\sigma\) falls within the range of \(10^{1}\) to \(10^{5}\) across all values of \(\sigma\). Additionally, in Fig. \ref{1a}, the proposed TMvRKM model attains its highest performance when \(\sigma\) is within \(10^{-5}\) to \(10^{1}\). Based on these observations, we suggest selecting \(\eta_1\) and \(\sigma\) within these ranges to achieve optimal results. However, fine-tuning may still be required depending on the specific dataset to ensure the best generalization performance for the proposed TMvRKM model.

\section{Conclusions}\label{Conclusions-section}
In this paper, we proposed multiview twin restricted kernel machine (TMvRKM), a novel model designed to address the challenges of multi-view learning (MVL). Our approach overcomes the limitations of traditional SVM-based MVL methods, particularly the computational inefficiencies and difficulties in handling view-inconsistent samples. The TMvRKM model leverages the kernel trick to project data into a high-dimensional feature space, where it identifies an optimal separating hyperplane through a regularized least squares approach. This allows for enhanced classification performance and computational efficiency, making the model highly scalable for large-scale datasets. The TMvRKM model distinguishes itself by efficiently incorporating both early and late fusion strategies, enabling it to capture complementary information from multiple views while maintaining flexibility for individual view variations. Experimental results on various benchmark datasets, including UCI, KEEL, and AwA, demonstrate the superior generalization performance of TMvRKM against the existing models. These findings affirm the effectiveness of the proposed model in addressing the computational and generalization challenges typically encountered in multi-view learning tasks. Furthermore, statistical analyses including ranking, the Friedman test, and the Nemenyi post hoc test confirm the significantly enhanced robustness of our proposed model comparison to existing models. Future work will focus on further enhancing the model’s capabilities and exploring its applicability in other domains, such as clustering and dimensionality reduction, which could reveal new insights and broaden its utility.

\bibliographystyle{IEEEtranN}
\bibliography{refs.bib}
\end{document}